\PassOptionsToPackage{table,dvipsnames}{xcolor}
\documentclass[letterpaper]{article} 
\usepackage{aaai2027} 
\usepackage[hyphens]{url} 
\usepackage{graphicx} 
\usepackage{natbib} 
\usepackage{comment}
\usepackage{caption} 
\usepackage{amsmath,amssymb}
\usepackage{booktabs}
\usepackage{makecell}
\usepackage{multirow}
\usepackage{array}
\usepackage{pifont}
\usepackage{algorithm}
\usepackage{algorithmic}

\newcommand{\Method}{{RoboReact}}

\definecolor{RRLeft}{HTML}{000000}
\definecolor{RRRight}{HTML}{000000}
\definecolor{RRBoth}{HTML}{000000}
\definecolor{RRStageOne}{HTML}{000000}
\definecolor{RRStageTwo}{HTML}{000000}
\definecolor{RRAvg}{HTML}{EDEDED}
\newcommand{\rrtight}[1]{%
    \begingroup\renewcommand{\arraystretch}{0.72}%
    \begin{tabular}[c]{@{}c@{}}#1\end{tabular}\endgroup}
\newcommand{\stepLeft}[1]{\textcolor{RRLeft}{\rrtight{#1}}}
\newcommand{\stepRight}[1]{\textcolor{RRRight}{\rrtight{#1}}}
\newcommand{\stepBoth}[1]{\textcolor{RRBoth}{\rrtight{#1}}}
\newcommand{\stepStageOne}[1]{\textcolor{RRStageOne}{\rrtight{#1}}}
\newcommand{\stepStageTwo}[1]{\textcolor{RRStageTwo}{#1}}
\newcommand{\rrgroupa}[1]{#1}
\newcommand{\rrgroupb}[1]{#1}
\newcolumntype{Y}{>{\columncolor{RRAvg}[1pt][1pt]\color{black}}c}
\newcommand{\rravgbox}[1]{{\setlength{\fboxsep}{1.3pt}\colorbox{RRAvg}{\color{black}#1}}}

\nocopyright

\begin{document}

\title{RoboReact: Agentic Skill Distillation from Generated Egocentric Videos for Generalizable Whole-Body Manipulation}

\author{
    Shuliang He\textsuperscript{\rm 1,2},
    Shuai Wang\textsuperscript{\rm 2},
    Bo Yue\textsuperscript{\rm 1},
    Junchi Teng\textsuperscript{\rm 2,3},
    Changyu Wang\textsuperscript{\rm 2},
    Guiliang Liu\textsuperscript{\rm 1}\thanks{Corresponding author}
}
\affiliations{
    \textsuperscript{\rm 1}The Chinese University of Hong Kong, Shenzhen\\
    \textsuperscript{\rm 2}JD Technology\\
    \textsuperscript{\rm 3}Tsinghua University\\
    sereinheshuliang@163.com, wangshuai.shawn@jd.com, boyue@link.cuhk.edu.cn, tjc21@mails.tsinghua.edu.cn, wangchangyu7@jd.com, liuguiliang@cuhk.edu.cn
}

\maketitle

\begin{abstract}
Humanoid robots have the potential to perform dexterous manipulation in human environments, yet acquiring diverse and generalizable skills remains costly due to expensive hardware data collection and labor-intensive annotation. 
Recent advances in video generative models provide a promising opportunity to synthesize rich manipulation experiences from visual observations, but transferring such imagined behaviors into executable whole-body humanoid skills remains largely unexplored. In this work, we present \Method{}, a framework that automatically synthesizes whole-body humanoid manipulation skills from a single egocentric RGB-D observation. \Method{} generates human manipulation videos, extracts geometry-preserving interaction keyframes through depth-aware 3D reconstruction, and retargets them to high-DoF humanoid platforms while preserving hand-object interaction geometry. To bridge the gap between imagined plans and physical execution, \Method{} performs online object-centric re-grounding and leverages a vision-language model-guided refinement loop to adapt skills under geometric mismatch and execution deviations. The refined skills are executed through a whole-body controller, enabling coordinated whole-body manipulation and dexterous interaction. Experiments on real humanoid robots demonstrate that \Method{} generalizes across diverse object configurations and robustly recovers from execution disturbances without requiring teleoperation or human demonstrations. These results highlight the potential of combining generative models, vision-language reasoning, and closed-loop control for scalable humanoid skill acquisition.
Project website: \url{https://roboreact.github.io/}.
\end{abstract}

\section{Introduction}
While humanoid robots hold great promise for performing dexterous manipulation in human environments, scaling such skills remains a fundamental challenge. 
Existing approaches typically rely on teleoperated demonstrations~\cite{ze2025twist,ben2024homie}, human-motion retargeting~\cite{yuan2025motiontrans,yang2025omniretarget}, or reinforcement learning-based fine-tuning~\cite{ankile2025residual,lin2025sim}. 
However, these approaches are often computationally expensive, labor-intensive, and tightly coupled to specific robot embodiments.

Recent advances in video generative models~\cite{wan2025wan,kong2024hunyuanvideo} provide a new opportunity to synthesize humanoid manipulation skills at scale. 
Given only a single RGB-D observation from the robot's egocentric view, generative models can imagine human-performed task executions and provide dense visual guidance for skill synthesis~\cite{du2023learning,bharadhwaj2024gen2act}. 
Compared with hardware-based data collection, this process can be largely automated, enabling scalable generation of diverse manipulation experiences across tasks and embodiments.

\begin{figure}[t]
    \centering
    \includegraphics[width=\columnwidth]{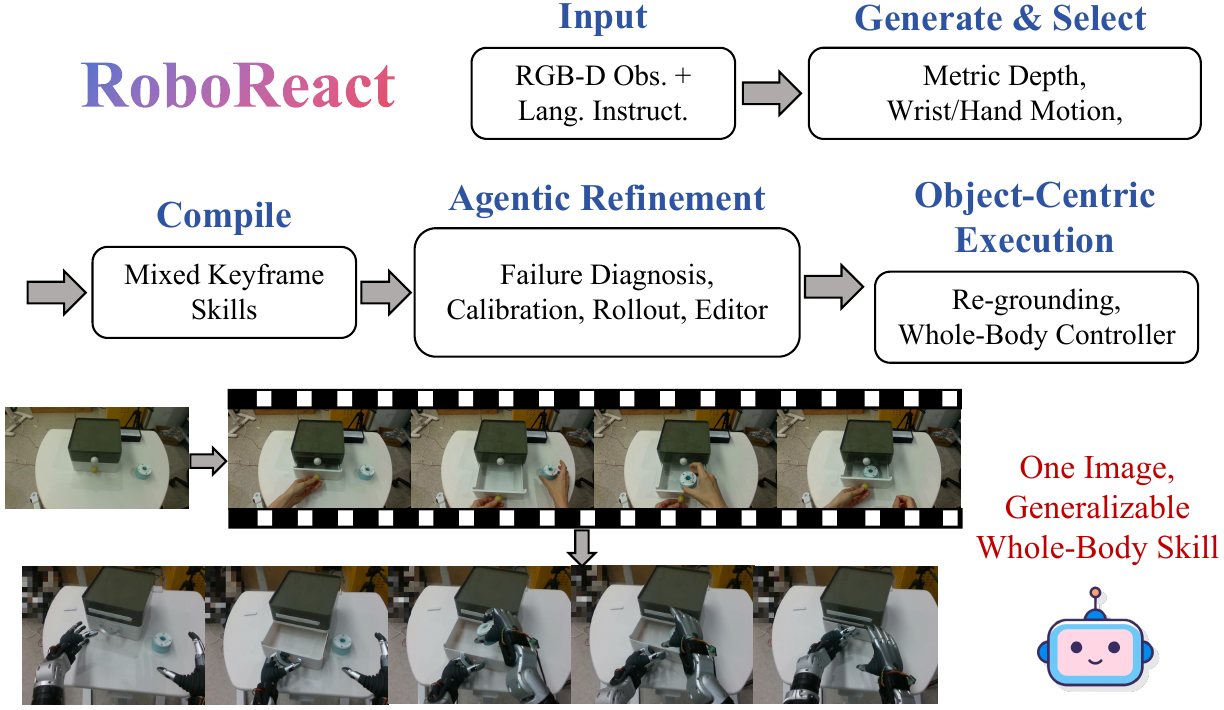}
    \caption{RoboReact distills a generated interaction video into an object-centric whole-body manipulation skill.}
    \label{fig:teaser}
\end{figure}

While egocentric visual guidance can offer generalizable cues for robotic manipulation, it typically provides only coarse-grained supervision (such as point trajectory~\cite{bharadhwaj2024track2act}, affordance~\cite{li2025affgrasp}, and 3D hand pose~\cite{kareer2024egomimicscalingimitationlearning}) over continuous hand trajectories. To enable reliable deployment on real robots, additional 3D spatial information~\cite{DBLP/cvpr/ChenSZPL25}, hand-object calibration~\cite{singh2024handobjectinteractionpretrainingvideos}, and real robot data finetuning~\cite{kareer2024egomimicscalingimitationlearning,singh2024handobjectinteractionpretrainingvideos} are often required to translate these visual signals into physically executable motions.
More critically, how to transfer such video-based guidance from low-DoF manipulation systems with fixed-base grippers to highly dynamic humanoid platforms~\cite{bharadhwaj2024track2act,DBLP/cvpr/ChenSZPL25,kareer2024egomimicscalingimitationlearning}, where whole-body coordination and loco-manipulation are essential, remains an open and largely unexplored challenge.

To address this challenge, we propose \Method{} to automatically synthesize executable whole-body humanoid manipulation skills (Figure~\ref{fig:teaser}). Specifically, given a single egocentric RGB-D frame, \Method{} generates a human manipulation video and extracts geometry-preserving interaction keyframes through depth-aware 3D reconstruction~\cite{wang2026vggtomega}. Rather than directly warping continuous trajectories, \Method{} preserves the underlying hand-object interaction geometry by retargeting human motion in keyframes to a high-DoF humanoid platform. However, directly executing such generated motions remains challenging: generative videos provide only an approximate estimate of the underlying interaction geometry, which may not accurately match the true object configuration and can lead to failures under perception uncertainty and real-world disturbances.

Striving for reliable humanoid manipulation skill, \Method{} closes the perception-to-action gap through online re-grounding, aligning the retargeted motion with the observed object pose, and leveraging a VLM-based trial-and-error loop to iteratively refine interactions and recover from execution deviations.
The refined skills are finally executed through a whole-body controller, enabling coordinated whole-body manipulation. Across four long-horizon real-world tasks, \Method{} outperforms ReKep and YOTO, achieving an average terminal success rate of $81.3\%$, without task-specific teleoperation or human demonstrations.
\begin{figure*}[t]
    \centering
    \includegraphics[width=\textwidth]{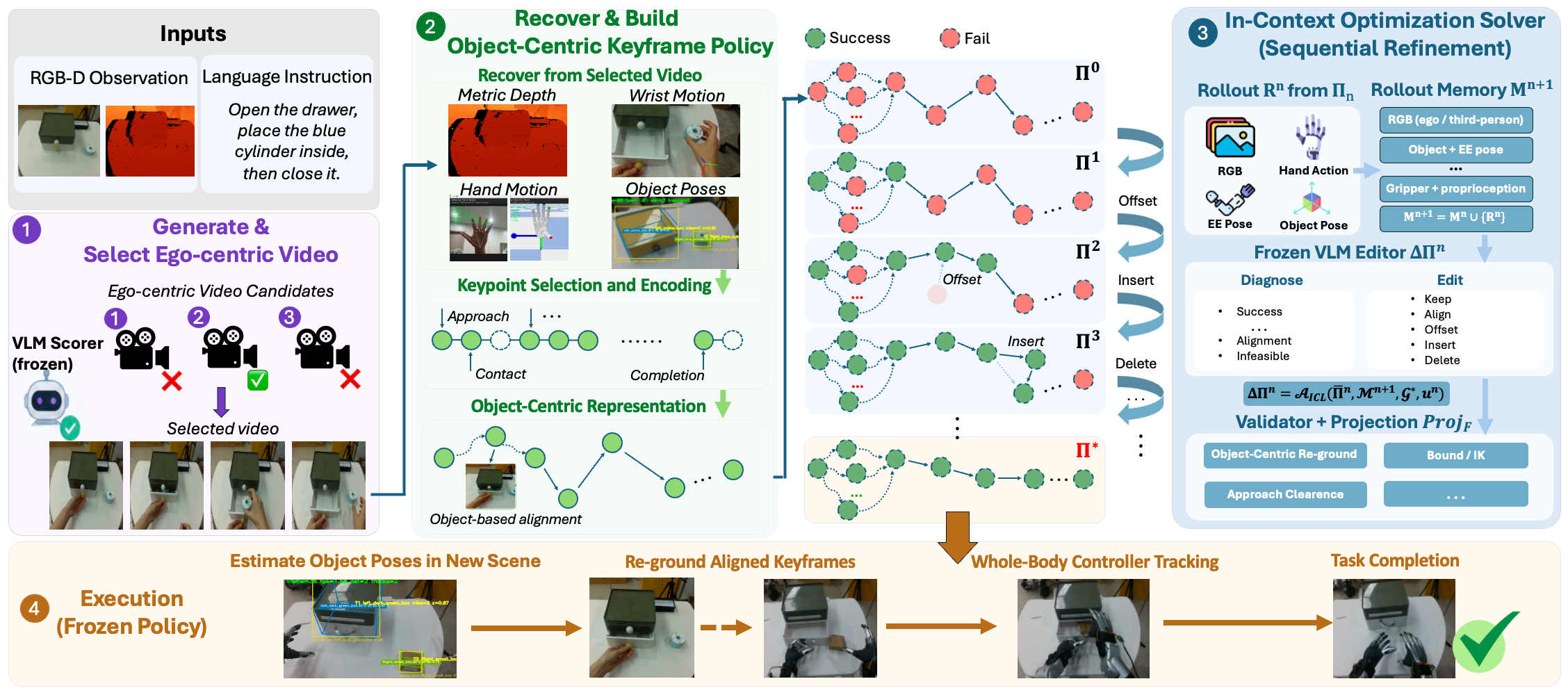}
    \caption{\textbf{Overview of \Method{}.} A VLM-selected generated video is compiled into an object-centric keyframe policy, refined from calibration rollouts, and executed through online object-pose re-grounding without test-time VLM access.}
    \label{fig:pipeline}
\end{figure*}

We summarize our main contributions as follows:
\begin{itemize}
\item RoboReact is the first framework to solve long-horizon,
generalizable whole-body manipulation using only pretrained models and a single
RGB-D frame as data source, without any teleoperated or human demonstrations.

\item RoboReact introduces an object-centric online re-grounding mechanism that uses a VLM-driven trial-and-error loop to iteratively align and refine retargeted whole-body skills, enabling robust generalization across variations in object shape, pose, and scene background.

\item RoboReact empirically demonstrates favorable scaling behavior: its performance consistently improves with more capable vision-language and video generation models, suggesting a clear path toward continued gains as foundation models advance.

\end{itemize}

\section{Related Works}
Human videos provide a scalable source of supervision for learning complex manipulation behaviors~\cite{grauman2024ego, hoque2025egodex}. To mitigate the embodiment gap, prior work leverages intermediate representations such as keypoints and hand poses~\cite{yang2025gripperkeypose, DBLP/cvpr/ChenSZPL25}, affordances~\cite{ma2025gloverpp, li2025affgrasp}, and object-centric flows~\cite{yin2025motionfield, li2025novaflow} to enable motion retargeting~\cite{yuan2025motiontrans, li2024okami} and action planning~\cite{kerr2024robot, chen2024object}. More recently, generative models treat \emph{AI-generated video} as an actionable plan, synthesizing future interaction sequences that are converted into robot actions~\cite{du2023learning, bharadhwaj2024gen2act}; we adopt this paradigm to guide dexterous whole-body interaction while grounding it in real-world geometry. Unlike \citet{patel2025rigvid}, which imitates a generated
video open-loop on a fixed-base arm, we re-ground the plan against the observed
object pose and refine it online, yielding stronger generalization to novel
object configurations and disturbances. The closest to our work is YOTO~\cite{zhou2025you}. While YOTO demonstrates bimanual learning from binocular video, it is limited by implicit depth, a fixed base, and low-DoF grippers; in contrast, we leverage RGB-D sensing for mobile, high-DoF dexterous retargeting. 

\textbf{Agent-based Manipulation.}
Leveraging LLM and VLM agents for robotics is an active area of research. Prior work uses agents for long-horizon task decomposition via grounded affordances~\cite{ahn2022saycan}, code-as-policy generation~\cite{liang2023code}, or symbolic planning~\cite{chen2025lammap}, and as spatial reasoners that translate intent into geometric grounding through 3D value maps~\cite{huang2023voxposer}, relational keypoint constraints~\cite{huang2024rekep}, or object-centric primitives~\cite{pan2025omnimanip}. Recent pipelines further extend to closed-loop execution with multi-agent coordination~\cite{guo2024malmm, yang2025maniagent} and VLM-driven failure recovery~\cite{chen2025robot}. However, these efforts operate on a fixed base over tabletop workspaces, reasoning about a decoupled arm without whole-body coordination or mobility; loco-manipulation agents~\cite{qiu2024wildlma, li2026w1} add mobility but typically dispatch pre-trained skill primitives, leaving fine-grained dexterous interaction outside the agent's loop. In contrast, our framework couples agent-based reasoning with mobile, whole-body dexterous execution, grounding each stage in interaction geometry retargeted from human video and re-invoking object-centric re-grounding and interference recovery online.

\section{Method}

\subsection{Problem Overview}

RoboReact is formulated as a constrained sequential skill-distillation problem.
Given a task prompt $P_{\mathrm{task}}$, the robot's initial RGB-D observation
$(I_1, D_1)$, and access to a pretrained video generation model $\mathcal{V}$,
our objective is to recover an executable and generalizable whole-body skill that
preserves the task semantics and hand--object interaction structure while
satisfying the constraints of the robot embodiment. The video generation model
serves as a prior over human interaction: conditioned on the task prompt and the
initial RGB frame, it synthesizes a human interaction video
$\mathcal{G} = \mathcal{V}(I_1, P_{\mathrm{task}})$, from
which the interaction structure is distilled. Unlike reinforcement learning,
RoboReact does not optimize a reward through gradient-based parameter updates.
Instead, it compiles $\mathcal{G}$ into a structured keyframe skill and uses a
frozen vision-language model (VLM) as an in-context optimization agent that
diagnoses calibration rollouts and proposes bounded, interpretable skill refinements.

The procedure has two phases. \emph{Skill distillation} generates a video prior, recovers metric wrist and hand motion, compiles keyframes, and refines the skill via calibration rollouts. At \emph{test time}, the skill is frozen: perception only re-estimates object poses and re-grounds keyframes before sending feasible commands to the whole-body controller, keeping the VLM out of the control loop.

\subsection{Constrained Skill Distillation Formulation}
We represent a bimanual robot skill as an ordered sequence of $K$ keyframes,
\begin{equation}
\Pi=\{(\rho_k,o_k,T^l_{a,k},T^r_{a,k},h^l_k,h^r_k,m_k)\}_{k=1}^{K},
\label{eq:skill-representation}
\end{equation}
where $\rho_k\in\{\mathrm{approach},\mathrm{align},\mathrm{fixed}\}$ specifies the stage of keyframe $k$: $\mathrm{approach}$ denotes the pre-contact stage, where the robot arm moves towards the object; $\mathrm{align}$ indicates the interaction stage, where the robot hand interacts with the object (e.g. grasp a cup); and $\mathrm{fixed}$ denotes a stage where the relative geometry between the end-effector (EE) and the object remains unchanged.
$o_k$ denotes the reference object, and $T^\diamond_{a,k}\in\mathrm{SE}(3)$ denotes the EE pose of arm $\diamond$, where the subscript $a$ indicates the arm EE.
Unless otherwise stated, all object and EE poses are expressed in the robot-base frame.
$h^\diamond_k$ is the dexterous-hand motion for hand $\diamond\in\{l,r\}$, and $m_k=(m^l_k,m^r_k)\in\{0,1\}^2$ is the hand validity mask. Specifically, $m^\diamond_k=1$ for a directly detected hand pose and $m^\diamond_k=0$ when detection fails and requires pose imputation. Missing EE poses, or hand motions before the first valid detection, fall back to the first subsequent valid observation, whereas later missing poses hold the most recent valid observation.

We formulate skill distillation from the video $\mathcal{G}$ as the following constrained objective:
\begin{align}
&\Pi^\star=\arg\min_{\Pi\in\mathcal{F}}\mathcal{L}(\Pi,\mathcal{G}),\label{eq:constrained-distillation}\\
&\text{where}\;\mathcal{L}=\lambda_s\mathcal{L}_{\mathrm{sem}}+
\lambda_g\mathcal{L}_{\mathrm{geo}}+
\lambda_m\mathcal{L}_{\mathrm{mot}},\nonumber
\end{align}
Here, $\mathcal{F}$ denotes the feasible robot-skill space, while $\lambda_s$, $\lambda_g$, and $\lambda_m$ weight the semantic, geometric, and motion-preservation objectives, respectively. Equation~(\ref{eq:constrained-distillation}) provides the formal objective for the in-context optimization solver introduced in the next section. The individual objectives are defined as follows:

\subsubsection{Semantic Preservation Objective.}

\begin{figure}[t]
    \centering
    \includegraphics[width=\columnwidth]{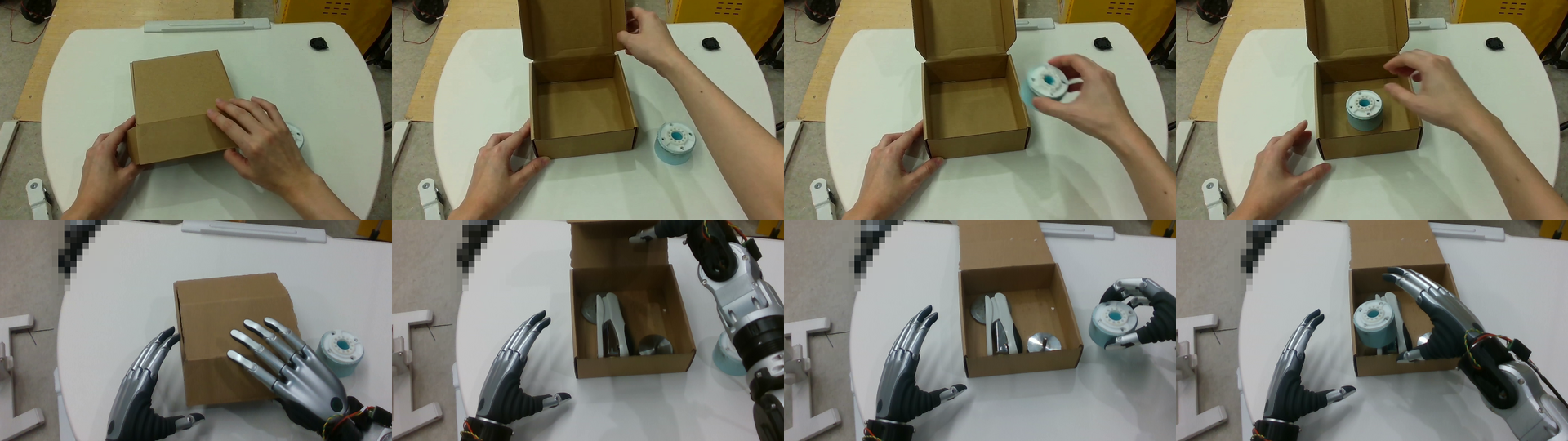}
    \caption{\textbf{Representative semantic keyframe comparisons.} The top row shows generated-video priors and the bottom row shows the corresponding real-robot rollout frames.}
    \label{fig:semantic-keyframe-comparison}
\end{figure}

The semantic term preserves the task objective, manipulation order, and interaction phases:
\begin{equation}
\mathcal{L}_{\mathrm{sem}}=d_{\mathrm{VLM}}\!\left(\phi(\mathcal{T}),\phi(\mathcal{G})\right),
\label{eq:semantic-loss}
\end{equation}
where $\phi(\cdot)$ denotes a structured visual-semantic description for robot rollout trace video $\mathcal{T}$ and generated human video $\mathcal{G}$ and $d_{\mathrm{VLM}}$ denotes discrepancies assessed by the frozen VLM. To keep this assessment grounded, the VLM cannot synthesize arbitrary continuous actions. We uniformly sample the generated video at 10\,Hz. The VLM selects an ordered subset $\mathcal{K}=\{t_k\}_{k=1}^{K}$ from these observed frames to encode the complete manipulation process, including approach, contact or grasp, manipulation, and completion. Figure~\ref{fig:semantic-keyframe-comparison} shows representative generated-prior and real-rollout pairs used for this semantic comparison.

\subsubsection{Interaction Geometry Objective.}

Direct trajectory replay is inappropriate because the generated human hand, robot hand, viewpoint, and workspace differ. RoboReact instead preserves object-centric interaction geometry. Because the generated video is non-metric, the human relative transform is not directly observable. At each aligned keyframe, the visible hand--object relation provides a semantic initialization that the VLM refines into an executable relative transform $\Delta T^{\diamond,*}_k$.
The corresponding geometry term is
\begin{equation}
\mathcal{L}_{\mathrm{geo}}=
\sum_{k,\diamond}m^\diamond_k\,
d_{\mathrm{SE}(3)}\!\left(\Delta T^{\diamond,\mathrm{robot}}_k,
\Delta T^{\diamond,*}_k\right).
\label{eq:geometry-loss}
\end{equation}
At execution time, the current object pose $\hat{T}_{o_k}$ re-grounds the template as
$\hat{T}^\diamond_{a,k}=\hat{T}_{o_k}\Delta T^{\diamond,*}_k$.
To avoid collisions when reaching an aligned pose, the VLM marks two preceding keyframes as approach frames and assigns object-conditioned clearance biases toward the refined transform. Aligned keyframes are used by default before and during contact, while fixed keyframes retain robot-frame commands without a meaningful visible object anchor.

\subsubsection{Motion Prior Objective.}

Generated videos provide a soft prior over reaching direction, wrist orientation, finger configuration, bimanual coordination, and temporal progression:
\begin{equation}
\mathcal{L}_{\mathrm{mot}}=\sum_{k,\diamond}m^\diamond_k\!\left[
d_T\!\left(T^\diamond_{a,k},\tilde{T}^\diamond_{a,k}\right)
+\alpha d_h\!\left(h^\diamond_k,\tilde{h}^\diamond_k\right)\right].
\label{eq:motion-loss}
\end{equation}
Here, $\tilde{T}^\diamond_{a,k}$ and $\tilde{h}^\diamond_k$ denote the wrist-pose and hand-command priors recovered from the selected generated video and retargeted to the robot spaces; $d_T\equiv d_{\mathrm{SE}(3)}$ and $d_h$ measure pose and hand-command discrepancies, and $\alpha$ balances the two terms. To construct this prior, we use off-the-shelf video generation, VLM selection, and hand-pose estimation to synthesize a candidate video from $(P_{\mathrm{task}}, I_1)$ and retarget its wrist and finger motion to the robot.

\subsubsection{Robot Feasibility Constraints.}

Let $\mathcal{F}$ denote policies passing deterministic checks for valid task structure, fresh object poses, supported edits, and joint-limited IK. The projection $\mathrm{Proj}_{\mathcal{F}}$ compiles a candidate by re-grounding object-relative keyframes, applying approach clearances, and converting wrist targets into joint targets; on failure, playback is disabled and the candidate is returned for refinement. These checks run on the initial policy and repeat after each refinement or test-time re-grounding.

\subsection{In-Context Optimization Solver}

In practice, directly solving this constrained optimization problem with first-order methods is challenging, as skill refinement is inherently sequential and requires semantic reasoning over intermediate outcomes. Instead of gradient-based optimization, we employ in-context refinement, where an agent iteratively analyzes, adapts, and improves the skill representation while satisfying the feasibility constraints.
At round $n$, the robot executes the projected skill $\bar{\Pi}^{n}$ in a calibration rollout and records a keyframe-level trace $\mathcal{R}^{n}$ containing egocentric and third-person RGB-D observations, estimated object poses, commanded and realized EE poses, hand commands, and proprioception. The accumulated memory is
$\mathcal{M}^{n+1}=\mathcal{M}^{n}\cup\{\mathcal{R}^{n}\}$.
Using the current skill, rollout memory, generated-video reference, and an optional sparse human hint $u^n$ that only describes an observed rollout phenomenon without prescribing a policy edit, the frozen VLM proposes
\begin{equation}
\Delta\Pi^n=\mathcal{A}_{\mathrm{ICL}}
(\bar{\Pi}^{n},\mathcal{M}^{n+1},\mathcal{G}^{\star},u^n).
\label{eq:icl-agent}
\end{equation}
Here, $\mathcal{A}_{\mathrm{ICL}}$ denotes the frozen VLM-based policy editor that maps these contextual inputs to the structured edit $\Delta\Pi^n$.
Before editing, the VLM assigns each relevant keyframe one of the labels \{\texttt{success}, \texttt{alignment}, \texttt{grasp}, \texttt{contact}, \texttt{infeasible}\} and cites the rollout evidence supporting its diagnosis. Its output is restricted to the structured operations \texttt{keep}, \texttt{align}, \texttt{offset}, \texttt{insert}, and \texttt{delete}. These operations may change an object association or execution mode, apply bounded pose or hand corrections, add a pre-contact or stabilization keyframe, or remove a redundant keyframe.

The refinement update is
\begin{equation}
\bar{\Pi}^{n+1}=\mathrm{Proj}_{\mathcal{F}}\!\left(
\Pi^n\oplus\mathcal{A}_{\mathrm{ICL}}
(\bar{\Pi}^{n},\mathcal{M}^{n+1},\mathcal{G}^{\star},u^n)\right),
\label{eq:skill-update}
\end{equation}
where $\oplus$ applies the structured edit. A deterministic validator checks the output schema, cited grounding, offset bounds, required keyframes, and preservation of previously successful subgoals. Invalid edits and edits whose projection fails are rolled back. Consequently, the VLM reasons about \emph{what} should change, whereas deterministic geometry and robot constraints decide whether the proposed change may be executed.

\subsection{Sequential Refinement Algorithm}

Algorithm~\ref{alg:sequential-refinement} formalizes the compile--refine--freeze procedure. The refinement state at round $n$ consists of the current feasible skill $\bar{\Pi}^{n}$ and accumulated rollout memory $\mathcal{M}^{n}$. Only validated edits can change the skill, so $\bar{\Pi}^{n}\in\mathcal{F}$ is maintained throughout the procedure.

\begin{algorithm}[t]
\caption{Sequential constrained skill distillation}
\label{alg:sequential-refinement}
\begin{algorithmic}[1]
\REQUIRE $P_{\mathrm{task}},(I_1,D_1)$; refinement budget $N_{\max}$
\ENSURE Frozen feasible skill $\bar{\Pi}^{\star}$
\STATE $\mathcal{G}^{\star}\leftarrow\mathrm{Select}(\mathrm{Generate}(P_{\mathrm{task}},I_1))$
\STATE $\bar{\Pi}^{0}\leftarrow\mathrm{Proj}_{\mathcal{F}}(\mathrm{Compile}(\mathcal{G}^{\star},I_1,D_1))$
\STATE $\mathcal{M}^{0}\leftarrow\emptyset$
\FOR{$n=0,\ldots,N_{\max}-1$}
    \STATE $\mathcal{R}^{n}\leftarrow\mathrm{Rollout}(\bar{\Pi}^{n})$;
    $\mathcal{M}^{n+1}\leftarrow\mathcal{M}^{n}\cup\{\mathcal{R}^{n}\}$
    \IF{$\psi(\mathcal{R}^{n})=1$}
        \RETURN $\bar{\Pi}^{n}$
    \ENDIF
    \STATE $\Delta\Pi^{n}\leftarrow\mathcal{A}_{\mathrm{ICL}}(\bar{\Pi}^{n},\mathcal{M}^{n+1},\mathcal{G}^{\star},u^n)$
    \IF{$\neg\mathrm{Valid}(\Delta\Pi^{n})$}
        \RETURN $\bar{\Pi}^{n}$
    \ENDIF
    \STATE $\bar{\Pi}^{n+1}\leftarrow\mathrm{Proj}_{\mathcal{F}}(\bar{\Pi}^{n}\oplus\Delta\Pi^{n})$
    \IF{projection fails}
        \STATE $\bar{\Pi}^{n+1}\leftarrow\bar{\Pi}^{n}$
    \ENDIF
\ENDFOR
\RETURN $\bar{\Pi}^{N_{\max}}$
\end{algorithmic}
\end{algorithm}

The low-level controller adopts HOMIE~\cite{ben2024homie}, written as $\pi^{\mathrm{low}}(a_t\mid q_{t-H},\ldots,q_t,\boldsymbol{c}_t,a_{t-1})$, where $\boldsymbol{c}_t$ contains the base, body-height, torso, arm, and hand targets obtained from the projected keyframe skill. RoboReact therefore defines and adapts the high-level interaction structure, while the low-level policy tracks feasible commands and maintains whole-body balance.
Overall, the unified agentic optimization here can be summarized as
\begin{equation}
\Pi^\star=\arg\min_{\Pi\in\mathcal{F}}\mathcal{L}(\Pi,\mathcal{G}),
\qquad
\Pi\leftarrow\mathcal{A}_{\mathrm{ICL}}(\Pi,\mathcal{M},\mathcal{G}),
\label{eq:unified-view}
\end{equation}
where generated videos provide the interaction prior and every agent-induced update is projected onto $\mathcal{F}$ via $\mathrm{Proj}_{\mathcal{F}}$.

\section{Experiments}
We organize the evaluation around four research questions.
\textit{Q1}: How effectively can \Method{} distill a generated interaction video into an executable whole-body skill?
\textit{Q2}: How do the calibration-rollout budget and the capability of the frozen VLM policy editor affect the quality of the distilled skill?
\textit{Q3}: How much do semantic keyframe selection, accumulated rollout memory, third-person RGB-D evidence, and video-generator quality each contribute?
\textit{Q4}: Once refinement ends and the VLM leaves the control loop, how much performance does object-centric execution retain under stage-specific perturbations?

\begin{table*}[t]
    \centering
    {
    \small
    \setlength{\tabcolsep}{0.8pt}
    \renewcommand{\arraystretch}{0.9}
    \begin{minipage}{\linewidth}
    \centering
    \begin{tabular*}{\linewidth}{@{\extracolsep{\fill}}lccccccYcccccY@{}}
        \toprule
        \multirow{2}{*}{\raisebox{-0.9ex}{Methods}} &
        \multicolumn{7}{c}{\texttt{Hand Over}} &
        \multicolumn{6}{c}{\texttt{Open Box}} \\
        \cmidrule(lr){2-8} \cmidrule(lr){9-14}
        & \stepRight{pick\\cup} & \stepRight{close to\\left hand} & \stepBoth{hand\\over} & \stepLeft{move to\\tray} & \stepLeft{place\\cup} & \makecell{\textbf{SR}\\\textbf{(\%)}} & \makecell{\textbf{Avg.}\\\textbf{Len.}}
        & \stepLeft{stabilize\\box} & \stepBoth{open\\box} & \stepRight{pick\\object} & \stepRight{place\\object} & \makecell{\textbf{SR}\\\textbf{(\%)}} & \makecell{\textbf{Avg.}\\\textbf{Len.}} \\
        \midrule
        ReKep & 12/20 & 10/20 & 9/20 & 7/20 & 7/20 & 35.0 & 2.25 & 8/20 & 4/20 & 13/20 & 3/20 & 15.0 & 1.40 \\
        YOTO & 16/20 & 16/20 & 15/20 & 15/20 & 15/20 & 75.0 & 3.85 & 17/20 & 14/20 & 18/20 & 13/20 & 65.0 & 3.10 \\
        One-Shot Real Prior & 19/20 & 18/20 & \textbf{18/20} & \textbf{17/20} & \textbf{17/20} & \textbf{85.0} & 4.45 & 18/20 & \textbf{16/20} & 18/20 & \textbf{14/20} & \textbf{70.0} & 3.30 \\
        \textbf{\Method{}} & \textbf{20/20} & \textbf{19/20} & 17/20 & \textbf{17/20} & \textbf{17/20} & \textbf{85.0} & \textbf{4.50} & \textbf{19/20} & 15/20 & \textbf{19/20} & \textbf{14/20} & \textbf{70.0} & \textbf{3.35} \\
        \midrule
    \end{tabular*}
    \begin{tabular*}{\linewidth}{@{\extracolsep{\fill}}lcccccccYcccccY@{}}
        \multirow{2}{*}{\raisebox{-0.9ex}{Methods}} &
        \multicolumn{8}{c}{\texttt{Pour Water}} &
        \multicolumn{6}{c}{\texttt{Open Drawer}} \\
        \cmidrule(lr){2-9} \cmidrule(lr){10-15}
        & \stepLeft{pick\\cup} & \stepRight{pick\\bottle} & \stepBoth{close\\by} & \stepRight{pour\\water} & \stepRight{place\\bottle} & \stepLeft{place\\cup} & \makecell{\textbf{SR}\\\textbf{(\%)}} & \makecell{\textbf{Avg.}\\\textbf{Len.}}
        & \stepLeft{open\\drawer} & \stepRight{pick\\object} & \stepRight{place\\object} & \stepLeft{close\\drawer} & \makecell{\textbf{SR}\\\textbf{(\%)}} & \makecell{\textbf{Avg.}\\\textbf{Len.}} \\
        \midrule
        ReKep & 13/20 & 13/20 & 10/20 & 9/20 & 8/20 & 10/20 & 40.0 & 3.15 & 6/20 & 11/20 & 4/20 & 5/20 & 20.0 & 1.30 \\
        YOTO & 18/20 & 17/20 & 17/20 & 17/20 & 16/20 & 16/20 & 80.0 & 5.05 & 16/20 & 18/20 & 15/20 & 16/20 & 75.0 & 3.25 \\
        One-Shot Real Prior & \textbf{20/20} & \textbf{18/20} & \textbf{18/20} & 16/20 & \textbf{18/20} & \textbf{18/20} & 80.0 & 5.40 & \textbf{18/20} & \textbf{20/20} & \textbf{18/20} & \textbf{17/20} & \textbf{85.0} & \textbf{3.65} \\
        \textbf{\Method{}} & \textbf{20/20} & \textbf{18/20} & \textbf{18/20} & \textbf{18/20} & 17/20 & \textbf{18/20} & \textbf{85.0} & \textbf{5.45} & 17/20 & 19/20 & 17/20 & \textbf{17/20} & \textbf{85.0} & 3.50 \\
        \bottomrule
    \end{tabular*}
    \end{minipage}
    }
    \caption{Quantitative comparisons across four long-horizon whole-body manipulation tasks. We report step-wise success rates, task success rates (SR), and the average completed task length (Avg. Len.). Bold entries denote the best-performing method.}
    \label{tab:roboreact-main-comparison}
\end{table*}

\subsection{Experiment Setups}
\textbf{Hardware.} All real-world experiments use a 29-DoF Unitree G1 equipped with two BrainCo Revo2 Touch dexterous hands. A head-mounted RealSense D435i provides egocentric RGB-D observations, while an external D435 records third-person calibration rollouts. WildDet3D~\cite{huang2026wilddet3d} provides online object-pose estimates, and an offboard workstation with an RTX 4080 Super runs perception and high-level control; the resulting commands are tracked by a HOMIE-based~\cite{ben2024homie} whole-body controller.

\textbf{Tasks.} We evaluate \Method{} on four long-horizon whole-body manipulation tasks spanning diverse bimanual coordination and contact patterns: \texttt{Hand Over}, \texttt{Pour Water}, \texttt{Open Box}, and \texttt{Open Drawer}. Here, L, R, and B denote left-hand, right-hand, and bimanual execution, respectively.
\texttt{Hand Over}: pick cup (R), close to left hand (R), hand over (B), move to tray (L), and place cup (L).
\texttt{Pour Water}: pick cup (L), pick bottle (R), close to each other (B), pour water (R), place bottle (R), and place cup (L).
\texttt{Open Box}: stabilize box (L), open box (B), pick object (R), and place object (R).
\texttt{Open Drawer}: open drawer (L), pick object (R), place object (R), and close drawer (L).

\textbf{Evaluation Protocol.}
Calibration and evaluation configurations are sampled independently. Object poses and scene backgrounds are randomized across trials, with successive poses differing by at least 5\,cm in translation and $10^\circ$ in rotation about one axis. \texttt{Open Box} and \texttt{Open Drawer} each use five unseen object instances of different shapes and sizes. Each task uses one frozen skill across all test configurations. To mitigate perceptual ambiguity in occluded calibration rollouts, each skill-distillation run permits at most five sparse human hints, each limited to a natural-language description of an observable failure without policy-edit command. No human input is permitted during test-time evaluation.

\begin{table*}[t]
    \centering
    {
    \small
    \setlength{\tabcolsep}{2.4pt}
    \renewcommand{\arraystretch}{0.9}
    \begin{tabular}{lcccccY ccccccY}
        \toprule
        \multirow{2}{*}{\makecell{Rounds}} &
        \multicolumn{6}{c}{\rrgroupa{\texttt{Hand Over}}} &
        \multicolumn{7}{c}{\rrgroupb{\texttt{Pour Water}}} \\
        \cmidrule(lr){2-7} \cmidrule(lr){8-14}
        & \stepRight{pick\\cup} & \stepRight{close to\\left hand} & \stepBoth{hand\\over} & \stepLeft{move to\\tray} & \stepLeft{place\\cup} & \textbf{Len.}
        & \stepLeft{pick\\cup} & \stepRight{pick\\bottle} & \stepBoth{close\\by} & \stepRight{pour\\water} & \stepRight{place\\bottle} & \stepLeft{place\\cup} & \textbf{Len.} \\
        \midrule
        0 & 1/13 & 0/13 & 0/13 & 0/13 & 0/13 & 0.08 & 2/13 & 3/13 & 2/13 & 0/13 & 0/13 & 0/13 & 0.54 \\
        5 & 8/13 & 8/13 & 5/13 & 5/13 & 4/13 & 2.31 & 7/13 & 8/13 & 6/13 & 3/13 & 3/13 & 2/13 & 2.23 \\
        10 & 11/13 & 11/13 & 10/13 & 9/13 & 8/13 & 3.77 & 11/13 & 10/13 & 9/13 & 8/13 & 8/13 & 8/13 & 4.15 \\
        15 & \textbf{13/13} & \textbf{13/13} & \textbf{12/13} & \textbf{12/13} & \textbf{11/13} & \textbf{4.69} & \textbf{12/13} & \textbf{13/13} & \textbf{12/13} & \textbf{11/13} & \textbf{11/13} & \textbf{11/13} & \textbf{5.38} \\
        \bottomrule
    \end{tabular}
    }
    \vspace{-0.04in}
    \caption{Step-wise success versus refinement budget over the \texttt{Hand Over} and \texttt{Pour Water} tasks.}
    \label{tab:roboreact-refine-stepwise}
    \vspace{-0.04in}
\end{table*}

\textbf{Baselines.} We compare ReKep~\cite{huang2024rekep}, YOTO~\cite{zhou2025you}, and a one-shot real-human-video prior. ReKep plans closed-loop end-effector trajectories from DINOv2 keypoints and GPT-4o-generated geometric constraints, representing demonstration-free constraint planning. YOTO extracts coordinated bimanual keyframes from a single human demonstration; we evaluate its action-injection variant, which performed comparably to its full diffusion policy in our trials. Both system baselines use the same HOMIE controller. The real prior replaces our generated video with a task-specific recording while retaining the same compilation, refinement, and frozen execution pipeline, thereby isolating the effect of the prior source. \Method{} is our complete method, initialized from a VLM-selected generated candidate. All methods use the same task decomposition and are evaluated over 20 trials per task in Table~\ref{tab:roboreact-main-comparison}.

\subsubsection{Metrics.}
We report task success rate (SR), step-wise success rates, and average completed step length (Avg. Len.). SR is the percentage of trials in which all required steps succeed; Avg. Len. is the mean number of completed steps. Some steps can still be evaluated after an earlier step fails, so step-wise success rates are not monotonic.
Best results within each directly comparable setting are shown in bold, including ties.
For refinement analysis, we report aggregate task success rates on \texttt{Hand Over}, \texttt{Pour Water}, and \texttt{Open Box}; the video-generator comparison uses \texttt{Pour Water} and the four-phase \texttt{Open Drawer} task.
Unless otherwise noted in the refinement-budget studies, the policies evaluated in Tables~\ref{tab:roboreact-main-comparison}, \ref{tab:roboreact-pour-ablation}, \ref{tab:roboreact-video-generator}, and \ref{tab:roboreact-recovery} are obtained after 20 refinement rounds.

\subsection{Experimental Results}
\begin{figure}[t]
    \centering
    \includegraphics[width=\columnwidth]{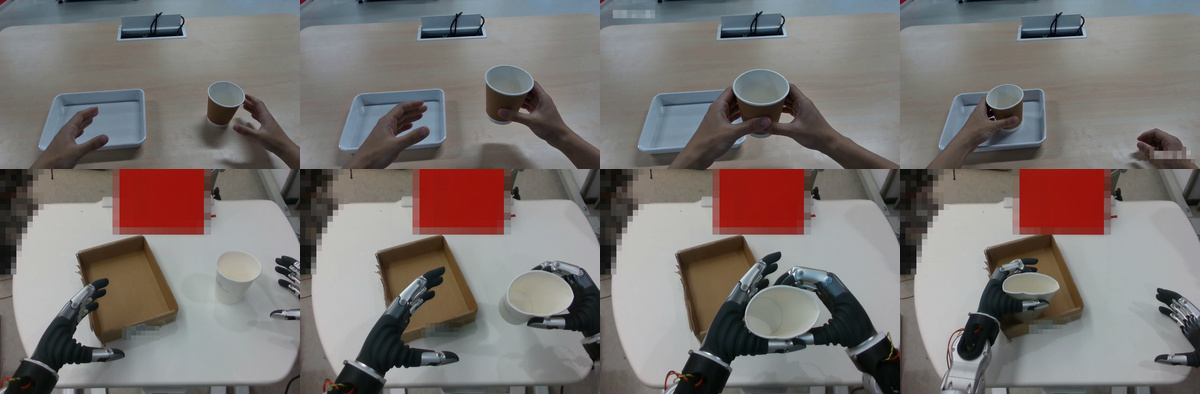}
    \caption{The required RGB-D source is not tied to a fixed acquisition setup.}
    \label{fig:q1-rgbd-source-diversity}
    \vspace{-0.1in}
\end{figure}

\begin{figure*}[t]
    \centering
    \includegraphics[width=\textwidth]{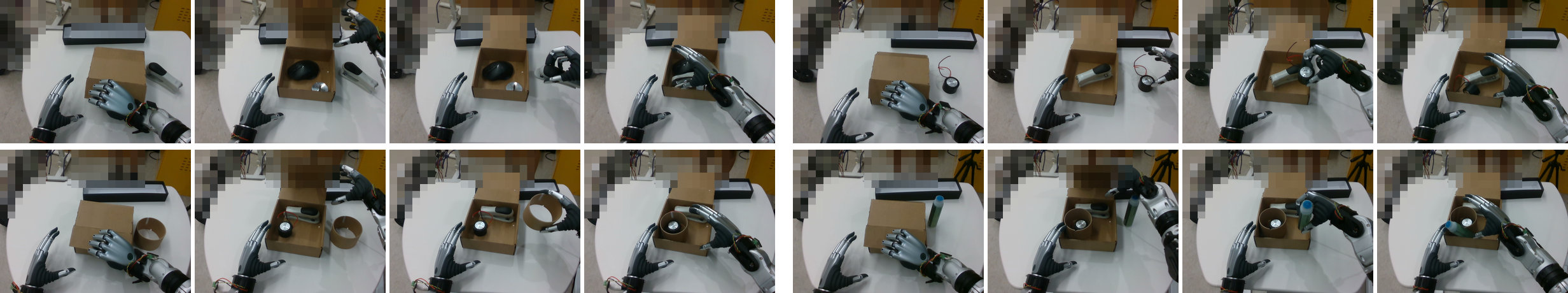}
    \caption{Each four-frame sequence shows contact with the lid, the fully opened box after releasing the lid, a stable object grasp, and release of the object inside the box.}
    \label{fig:q1-open-box-rollouts}
    \vspace{-0.1in}
\end{figure*}

\textbf{\textit{(Q1)} Effectiveness of generated-video skill distillation.}
Table~\ref{tab:roboreact-main-comparison} shows that \Method{} consistently outperforms ReKep and YOTO across all four tasks. \Method{} achieves nearly the same performance as the one-shot real-video prior: their mean SRs are $81.3\%$ and $80.0\%$, respectively, and both achieve a mean Avg. Len. of 4.20.
\Method{} achieves this performance without requiring task-specific human demonstration recordings, supporting our central design premise that the value of a non-metric generated video lies in its task order and hand--object interaction structure, which object-centric compilation and physical calibration make executable.
Figure~\ref{fig:q1-rgbd-source-diversity} qualitatively illustrates this transfer: the required RGB-D source observation can be acquired across different table heights, physical locations, and object instances or configurations, while the distilled interaction structure remains executable after object-centric re-grounding.
Figure~\ref{fig:q1-open-box-rollouts} further shows four successful \texttt{Open Box} executions with aligned interaction stages across different object poses, categories, and scene configurations.
The greater sensitivity of \texttt{Open Drawer} to the video-generator version, as shown in Table~\ref{tab:roboreact-video-generator}, suggests that contact-rich articulated interactions remain dependent on the fidelity of the upstream video prior.

\textbf{\textit{(Q2)} Effect of sequential refinement and editor capability.}
Table~\ref{tab:roboreact-refine-stepwise} shows non-decreasing success at every step-wise checkpoint across all refinement budgets; after 15 rounds, both tasks reach $11/13$ terminal completions. The gap between early-step and terminal-step success widens at 5 rounds but narrows at 10 and 15. This shift of the failure frontier toward task completion indicates that calibration rollouts improve downstream contact, manipulation, and release phases, not merely initial reachability.

\begin{table}[tbp]
    \centering
    {
    \small
    \setlength{\tabcolsep}{3.4pt}
    \renewcommand{\arraystretch}{0.95}
    \begin{tabular}{@{}llcYcYcY@{}}
        \toprule
        \multirow{2}{*}{Editor} & \multirow{2}{*}{Task} &
        \multicolumn{2}{c}{\rrgroupa{5}} & \multicolumn{2}{c}{\rrgroupa{10}} & \multicolumn{2}{c}{\rrgroupa{15}} \\
        \cmidrule(lr){3-4} \cmidrule(lr){5-6} \cmidrule(lr){7-8}
        & & SR & Len. & SR & Len. & SR & Len. \\
        \midrule
        \multirow{2}{*}{5.1-mini}
        & \texttt{Pour Water} & 15.4 & 1.77 & 38.5 & 3.15 & 69.2 & 4.54 \\
        & \texttt{Open Box} & 7.7 & 0.92 & 30.8 & 1.77 & 53.8 & 2.46 \\
        \midrule
        \multirow{2}{*}{\textbf{5.6-ultra}}
        & \texttt{Pour Water} & 15.4 & 2.23 & 61.5 & 4.15 & \textbf{84.6} & \textbf{5.38} \\
        & \texttt{Open Box} & 23.1 & 1.46 & 61.5 & 2.69 & \textbf{76.9} & \textbf{3.23} \\
        \bottomrule
    \end{tabular}
    }
    \caption{Task success rate and average completed step length under two GPT-Codex policy editors. Each editor-task entry contains 13 trials.}
    \label{tab:roboreact-refinement-editor}
    \vspace{-0.1in}
\end{table}

Table~\ref{tab:roboreact-refinement-editor} shows that the stronger editor exploits refinement rollouts more efficiently. After 10 rounds, it raises SR by 23.1 and 30.8 points on \texttt{Pour Water} and \texttt{Open Box}. With the VLM frozen, these gains come from converting in-context rollout evidence into structured, bounded edits rather than from parameter learning. This trend is consistent with GPT-5.6's reported gains on multimodal and visual-interactive benchmarks, including the 3D-spatial BenchCAD benchmark~\cite{openai2026gpt56,benchcad2026}, but does not isolate spatial perception alone.

\begin{table}[t]
    \centering
    {
    \small
    \setlength{\tabcolsep}{3.4pt}
    \renewcommand{\arraystretch}{1.0}
    \begin{tabular}{@{}lccccccY@{}}
        \toprule
        Variant & \stepLeft{pick\\cup} & \stepRight{pick\\bottle} & \stepBoth{close\\by} & \stepRight{pour\\water} & \stepRight{place\\bottle} & \stepLeft{place\\cup} & \textbf{Len.} \\
        \midrule
        w/o KS & 8/13 & 10/13 & 8/13 & 6/13 & 9/13 & 7/13 & 3.69 \\
        w/o Mem. & 10/13 & 9/13 & 9/13 & 8/13 & 7/13 & 8/13 & 3.92 \\
        w/o 3rd. & 11/13 & 12/13 & 11/13 & 6/13 & \textbf{12/13} & 10/13 & 4.77 \\
        \textbf{\Method{}} & \textbf{12/13} & \textbf{13/13} & \textbf{12/13} & \textbf{12/13} & \textbf{12/13} & \textbf{12/13} & \textbf{5.62} \\
        \bottomrule
    \end{tabular}
    }
    \caption{Step-wise ablation on \texttt{Pour Water}. KS denotes VLM-based semantic keyframe selection (replaced with uniform keyframes); Mem. denotes rollout memory for VLM policy refinement; 3rd. denotes the third-person camera.}
    \label{tab:roboreact-pour-ablation}
\end{table}

\textbf{\textit{(Q3)} Contribution of video-prior quality and refinement context.}
Table~\ref{tab:roboreact-video-generator} reveals a clear hierarchy among the complementary information sources. Seedance~1.5 Pro produces implausible hand scales, yielding a poorer interaction prior. Removing either semantic keyframe selection or rollout memory degrades Len. about twice as much as removing the third-person camera. These two support distinct aspects of long-horizon reasoning: semantic keyframes retain the critical interaction phases defining task progress, while rollout memory links the current failure to evidence from previous attempts. The third-person camera has a smaller aggregate effect but a highly localized one—removing it cuts pouring-phase success from $12/13$ to $6/13$—indicating the external view mainly resolves occluded spatial alignment and contact rather than improving perception uniformly.

\begin{table}[t]
    \centering
    {
    \small
    \setlength{\tabcolsep}{1mm}
    \renewcommand{\arraystretch}{1.05}
    \begin{tabular}{@{}lcccc@{}}
        \toprule
        \multirow{2}{*}{\makecell[l]{Video\\generator}} &
        \multicolumn{2}{c}{\rrgroupa{\texttt{Pour Water}}} &
        \multicolumn{2}{c}{\rrgroupb{\texttt{Open Drawer}}} \\
        \cmidrule(lr){2-3} \cmidrule(lr){4-5}
        & SR & \rravgbox{Len.} & SR & \rravgbox{Len.} \\
        \midrule
        Seedance 1.5 Pro & 84.6 & \rravgbox{5.23} & 69.2 & \rravgbox{3.00} \\
        \textbf{Seedance 2.0} & \textbf{92.3} & \rravgbox{\textbf{5.62}} & \textbf{84.6} & \rravgbox{\textbf{3.54}} \\
        \bottomrule
    \end{tabular}
    }
    \caption{Task success rate and average completed step length for video priors generated by Seedance 1.5 Pro and Seedance 2.0. Each task--generator entry contains 13 trials.}
    \label{tab:roboreact-video-generator}
    \vspace{-0.1in}
\end{table}

Table~\ref{tab:roboreact-video-generator} provides a complementary upstream analysis. Under the same downstream refinement procedure, a stronger video generator yields higher post-refinement performance, with an SR gain on \texttt{Open Drawer} that is twice that on \texttt{Pour Water}. This indicates that refinement does not eliminate differences in prior quality; instead, the two stages play complementary roles. The generated video supplies the task order and interaction structure, while rollout-grounded editing calibrates this prior into the robot’s embodiment and execution geometry.
\begin{table}[t]
    \centering
    {
    \small
    \setlength{\tabcolsep}{3.8pt}
    \renewcommand{\arraystretch}{1.0}
    \begin{tabular}{@{}lcccccccY@{}}
        \toprule
        \multirow{2}{*}{\raisebox{-1.4ex}{Case}} &
        \multirow{2}{*}{\raisebox{-1.4ex}{\stepStageOne{\\Squat}}} &
        \multicolumn{6}{c}{\stepStageTwo{\textbf{Manipulation}}} &
        \multirow{2}{*}{\textbf{Len.}} \\
        \cmidrule(lr){3-8}
        & & \stepLeft{pick\\cup} & \stepRight{pick\\bottle} & \stepBoth{close\\by} & \stepRight{pour\\water} & \stepRight{place\\bottle} & \stepLeft{place\\cup} & \\
        \midrule
        1 & 11/13 & 10/13 & 10/13 & 10/13 & 9/13 & 9/13 & 9/13 & 5.23 \\
        2 & 11/13 & 10/13 & 11/13 & 10/13 & 10/13 & 10/13 & 10/13 & 5.54 \\
        3 & \textbf{12/13} & \textbf{12/13} & 12/13 & 11/13 & 11/13 & 11/13 & 11/13 & 6.15 \\
        4 & \textbf{12/13} & \textbf{12/13} & \textbf{13/13} & \textbf{12/13} & \textbf{12/13} & \textbf{12/13} & \textbf{12/13} & \textbf{6.54} \\
        \bottomrule
    \end{tabular}
    }
    \caption{Step-wise recovery on \texttt{Pour Water} under two-stage disturbance cases. Squatting uses a table height randomly lowered by 0--25\,cm from nominal. Case 1: Squatting perturbation; Case 2: object shifted before manipulation; Case 3: base-pose perturbation at the operation position; Case 4: no disturbance.}
    \label{tab:roboreact-recovery}
\end{table}

\begin{figure}[t]
    \centering
    \includegraphics[width=\columnwidth]{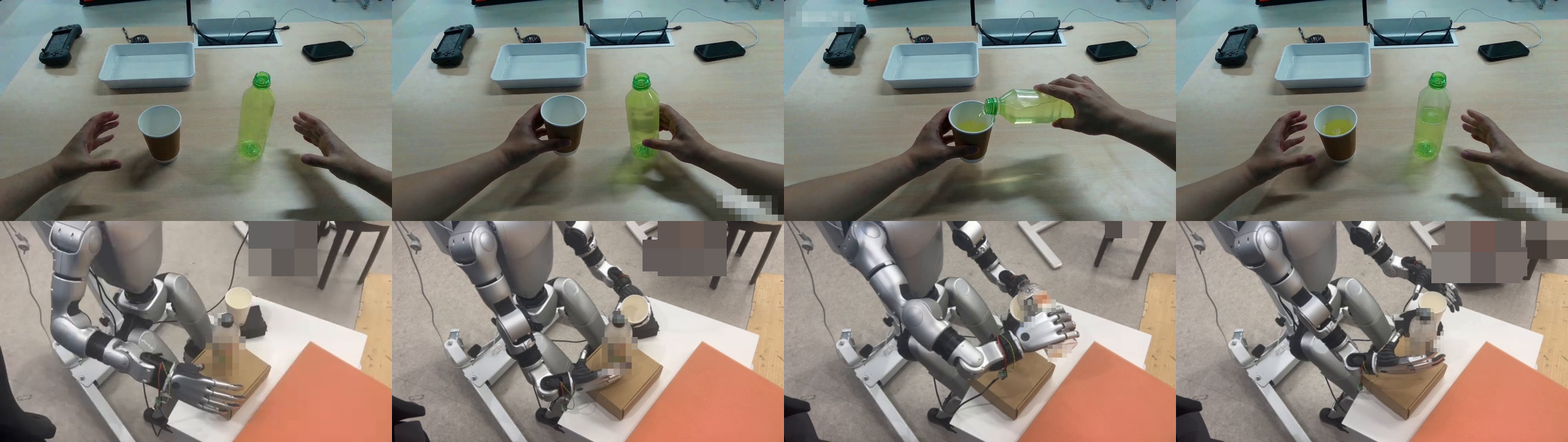}
    \caption{The robot first squats down to reach a manipulation-ready height and then performs the manipulation.}
    \label{fig:q4-pour-water-execution}
    \vspace{-0.1in}
\end{figure}

\textbf{\textit{(Q4)} Robustness of frozen object-centric execution.}
Table~\ref{tab:roboreact-recovery} shows that the frozen execution stack retains 80--94\% of nominal Avg. Len. under all tested disturbances, with terminal completion remaining above two-thirds even in the hardest case. Figure~\ref{fig:q4-pour-water-execution} illustrates the execution sequence: the robot first squats down to a manipulation-ready height and then performs the object-centric operation. More revealing than the absolute scores is their ordering. A base-pose perturbation at the operation position has the smallest effect, an object shift before manipulation has a larger effect, and perturbing squatting is most damaging. In the last case, squatting success itself drops by only one trial relative to nominal, yet terminal manipulation drops from $12/13$ to $9/13$. This indicates an error-propagation effect: an upstream posture deviation can leave the robot ready while degrading the reachable geometry of every subsequent contact. Conversely, the stronger performance under local object disturbances is consistent with the benefit of re-estimating object poses and re-grounding object-relative keyframes once the robot is already within a valid operating region. 
Since the VLM is absent at test time and no other individual recovery component is considered, Table~\ref{tab:roboreact-recovery} demonstrates the robustness of the execution stack.

\section{Conclusion}
In this paper, we introduced RoboReact, a framework that distills generalizable whole-body manipulation skills from a single egocentric RGB-D observation and a language instruction. RoboReact generates and selects a human interaction video, recovers metric wrist and hand motion, and compiles the interaction geometry into an object-centric keyframe skill. A frozen VLM then refines it from calibration-rollout evidence through bounded, structured edits, while deterministic feasibility projection blocks invalid commands. The refined skill is frozen and executed via online object-centric re-grounding and whole-body control, keeping the VLM out of the test-time loop. Across four long-horizon bimanual real-world tasks, RoboReact nearly matches a real-human-video prior, consistently outperforms structured and video-transfer baselines, and stays robust to object configurations and disturbances—pointing toward scalable humanoid skill acquisition through generative visual priors, grounded agentic refinement, and reliable whole-body execution.

\bibliography{references}

\end{document}